\documentclass[runningheads,a4paper]{llncs}

\usepackage{amssymb}
\usepackage{graphicx}

\usepackage{amsmath}
\usepackage{amssymb}
\usepackage{subfigure}
\usepackage{array}
\usepackage{multirow}
\usepackage{sidecap}
\usepackage{dsfont}

\usepackage{fancyvrb}

\usepackage{url}
\urldef{\mailoa}\path|ognjen.arandjelovic@gmail.com|

\graphicspath{{../imgs/}}

\def\Hline{\noalign{\hrule height 4\arrayrulewidth}}
\newcolumntype{V}{>{$\vcenter\bgroup\hbox\bgroup}c<{\egroup\egroup$}}

\begin{document}

\mainmatter

\title{Understanding Ancient Coin Images}


\author{Jessica Cooper \and Ognjen Arandjelovi\'c}
\authorrunning{Cooper and Arandjelovi\'c}
\institute{University of St Andrews, UK\\
\texttt{\{jessicamarycooper,ognjen.arandjelovic\}@gmail.com}}

\toctitle{~}
\tocauthor{Visually Understanding Rather Than Merely Matching Ancient Coin Images}
\maketitle

\begin{abstract}
In recent years, a range of problems within the broad umbrella of automatic, computer vision based analysis of ancient coins has been attracting an increasing amount of attention. Notwithstanding this research effort, the results achieved by the state of the art in the published literature remain poor and far from sufficiently well performing for any practical purpose. In the present paper we present a series of contributions which we believe will benefit the interested community. Firstly, we explain that the approach of visual matching of coins, universally adopted in all existing published papers on the topic, is not of practical interest because the number of ancient coin types exceeds by far the number of those types which have been imaged, be it in digital form (e.g.\ online) or otherwise (traditional film, in print, etc.). Rather, we argue that the focus should be on the understanding of the semantic content of coins. Hence, we describe a novel method which uses real-world multimodal input to extract and associate semantic concepts with the correct coin images and then using a novel convolutional neural network learn the appearance of these concepts. Empirical evidence on a real-world and by far the largest data set of ancient coins, we demonstrate highly promising results.
\end{abstract}

\section{Introduction}
Numismatics is the study of currency, including coins, paper money, and tokens. This discipline yields fascinating cultural and historical insights, and is a field of great interest to scholars, amateur collectors, and professional dealers alike. Important applications of machine learning in this field include theft detection, identification and classification of finds, and forgery prevention. The present work focuses on ancient coins.

Individual ancient coins of the same type can vary widely in appearance due to centering, wear, patination, and variance in artistic depiction of the same semantic elements. This poses a range of technical challenges and makes it difficult to reliably identify which concepts are depicted on a given coin using machine learning and computer vision based automatic techniques \cite{Aran2010}. For this reason, ancient coins are typically identified and classified by professional dealers or scholars, which is a time consuming process demanding years of experience due to the specialist knowledge required. Human experts attribute a coin based on its denomination, the ruler it was minted under and the time and place it was minted. A variety of different characteristics are commonly used for attribution, including:
\begin{itemize}
\item Physical characteristics such as weight, diameter, die alignment and colour,\\[-5pt]
\item Obverse legend, which includes the name of the issuer, titles or other designations,\\[-5pt]
\item Obverse motif, including the type of depiction (head or bust, conjunctional or facing, etc.), head adornments (bare, laureate, diademed, radiate, helmet, etc.), clothing (draperies, breastplates, robes, armour, etc.), and miscellaneous accessories (spears, shields, globes, etc.),\\[-5pt]
\item Reverse legend, often related to the reverse motif,\\[-5pt]
\item Reverse motif (primary interest herein), e.g.\ person (soldier, deity, etc.), place (harbour, fortress, etc.) or object (altar, wreath, animal, etc.), and\\[-5pt]
\item Type and location of any mint markings.
\end{itemize}

\subsection{Relevant prior work}
Most existing algorithmic approaches to coin identification are local feature based which results in poor performance due to loss of spatial relationships between elements \cite{Aran2010,RieuAran2015,RieuAran2016}. To overcome this limitation, approaches which divide a coin into segments have been proposed but these assume that coins are perfectly centred, accurately registered, and nearly circular in shape \cite{FareAran2017}. Recent work shows that existing approaches perform poorly on real world data because of the fragility of such assumptions \cite{FareAran2017}. Although the legend can be a valuable source of information, relying on it for identification is problematic because it is often significantly affected by wear, illumination, and minting flaws \cite{Aran2012d}. Lastly, numerous additional challenges emerge in and from the process of automatic data preparation, e.g.\ segmentation, normalization of scale, orientation, and colour \cite{ConnAran2017}. 

In the context of the present work it is particularly important to note that all existing work in this area is inherently limited by the reliance on visual matching \cite{ZahaKampZamb2007,KampZaha2008,Aran2010,AnwaZambKamp2013,AnwaZambKamp2015} and the assumption that the unknown query coin is one of a limited number of gallery types. However, this assumption is unrealistic as there are hundreds of thousands of different coin types \cite{Matt1966}. Hence herein our idea is to explore the possibility of \emph{understanding} the artistic content depicted on a coin, which could then be used for subsequent text based matching, allowing access to a far greater number of coin types without requiring them to be represented in the training data. In short, our ultimate aim is to attempt to a describe coin in the same way that a human expert would -- by identifying the individual semantic elements depicted thereon.

\section{Problem specification, constraints, and context}
Deep learning has proven successful in a range of image understanding tasks \cite{YueDimiAran2019}. Recent pioneering work on its use on images of coins has demonstrated extremely promising results, outperforming more traditional approaches by an order of magnitude \cite{SchlAran2017}. A major difference in the nature of the challenge we tackle here is that our data set is weakly supervised -- that is, samples are labelled only at the image level by a corresponding largely unstructured textual description, even though the actual semantic elements of interest themselves occupy a relative small area of the image. This kind of data is much more abundant and easier to obtain than fully supervised data (pixel level labels), but poses a far greater challenge.

\subsection{Challenge of weak supervision}
\begin{figure*}
\centering
\includegraphics[height=0.15\columnwidth]{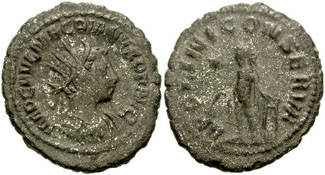}\hspace{15pt}
\includegraphics[height=0.15\columnwidth]{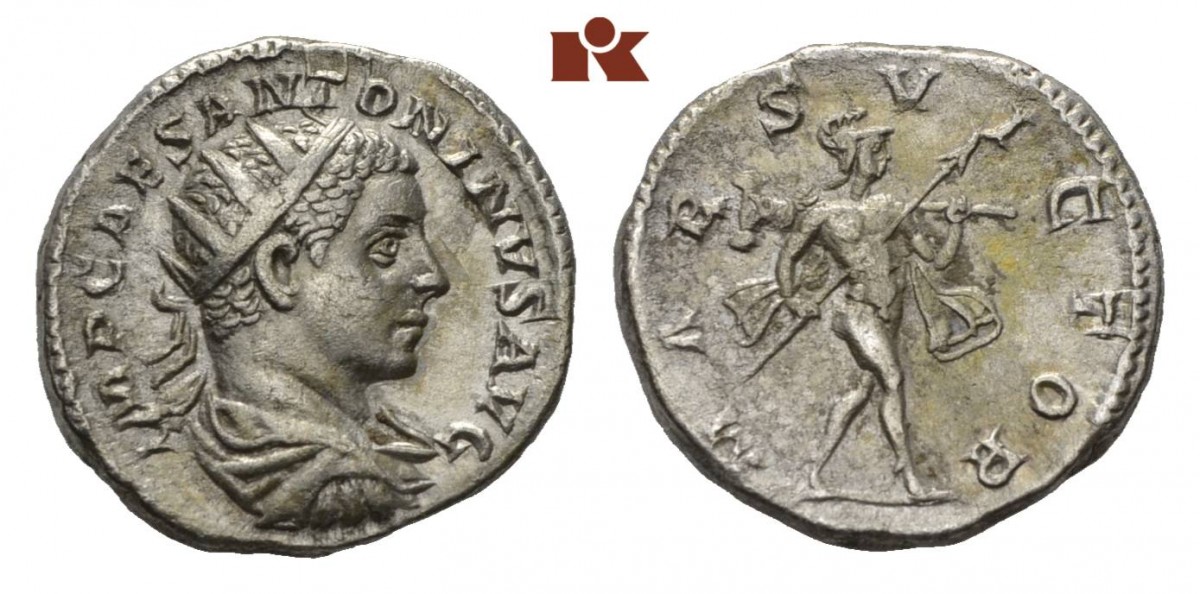}\hspace{15pt}
\includegraphics[height=0.15\columnwidth]{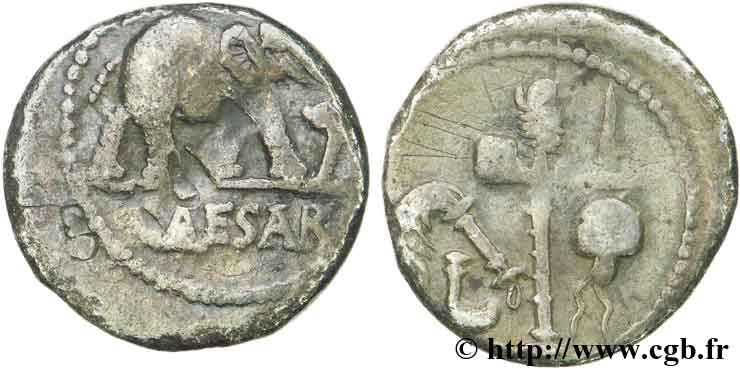}
\caption{Typical images used in the present work (obverses on the left, reverses on the right).}
\label{f:dataEx}
\end{figure*}

Recall that our overarching goal is to identify the presence of various important semantic elements on an unknown query coin's reverse. Moreover, we are after a scalable approach -- one which could be used with a large and automatically extracted list of elements. As this is the first attempt at solving the suggested task, we chose to use images which are relatively uncluttered, thus disentangling the separate problems of localizing the coin in an image and segmenting it out. Representative examples of images used can be seen in Fig~\ref{f:dataEx}.

To facilitate the automatic extraction of salient semantic elements and the learning of their visual appearance, each coin image is associated with an unstructured text description of the coin, as provided by the professional dealer selling the coin; see Fig~\ref{f:dataTxt}. The information provided in this textual meta-data varies substantially -- invariably it includes the descriptions of the coin's obverse and reverse, and the issuing authority (the person, usually emperor, on the obverse), but it may also feature catalogue references, minting date and place, provenance, etc.

\begin{figure}
\centering
\begin{Verbatim}[frame=single]
83, Lot: 226. Estimate \$100. Sold for \$92.

MACRIANUS. 260-261 AD. Antoninianus (22mm, 3.40 gm). Samosata? mint. Radiate, 
draped, and cuirassed bust  right / Apollo standing facing, head left, holding
laurel-branch and leaning on lyre; star in left  field. Cf. RIC V 6 (Antioch); 
MIR 44, 1728k; cf. Cohen 2 (same). VF, light porosity. Rare.
\end{Verbatim}
\vspace{-12pt}
\caption{Typical unstructured text attribution of a coin.}
\label{f:dataTxt}
\end{figure}

\begin{SCfigure}
\centering
\includegraphics[width=.4\columnwidth]{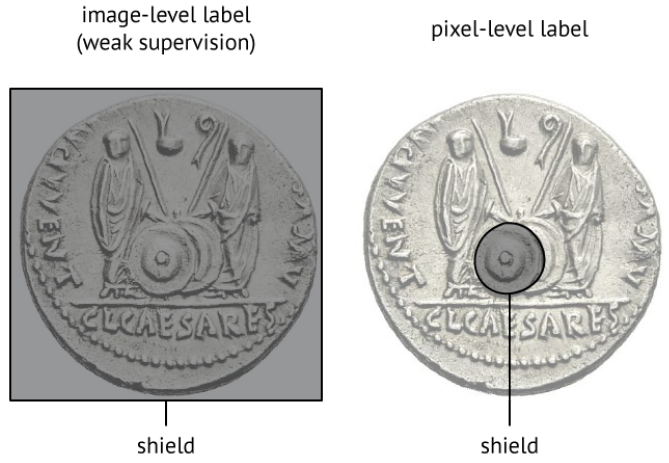}
\caption{The amount of labour which would be required to provide pixel-level labelling (right) of different concepts is prohibitive, imposing the much more challenging task of having to perform learning from weakly annotated data (left).}
\label{f:trainSupervision}
\end{SCfigure}

We have already emphasised the practical need for a scalable methodology. Consequently, due to the nature of the available data (which itself is representative of the type of data which can be readily acquired in practice), at best it is possible to say whether a specific semantic element (as inferred from text; see next section) is present in an image or not, see Fig~\ref{f:trainSupervision}. Due to the amount of human labour required it is not possible to perform finer labelling i.e.\ to specify where the element is, or its shape (there are far too many possible elements and the amount of data which would require labelling is excessive), necessitating weak supervision of the sort illustrated in Fig~\ref{f:trainSupervision}.

\subsection{Data pre-processing and clean-up}
As in most real-world applications unstructured, loosely standardized, and heterogeneous data, a major challenge in the broad domain of interest in the present work is posed by possibly erroneous labelling, idiosyncrasies, etc. Hence, robust pre-processing and clean-up of data is a crucial step for facilitating learning.

\subsubsection{Image based pre-processing}
We begin by cropping all images to the square bounding box which contains just the reverse of the corresponding coin, and isotropicaly resizing the result to the uniform scale of  $300 \times 300$ pixels. The obverse of a coin typically depicts the bust of an emperor, whilst the reverse contains the semantic elements of interest herein. 

\subsubsection{Text based extraction of semantics}
\begin{figure*}
\centering
\includegraphics[width=0.9\columnwidth]{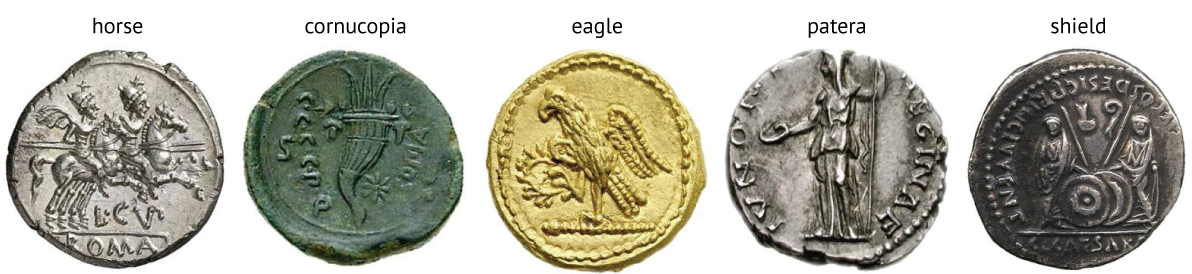}
\caption{Examples of depictions of reverse motif elements the present work focuses on.}
\label{f:elements}
\end{figure*}

\begin{SCfigure}
\centering
\includegraphics[width=0.65\columnwidth]{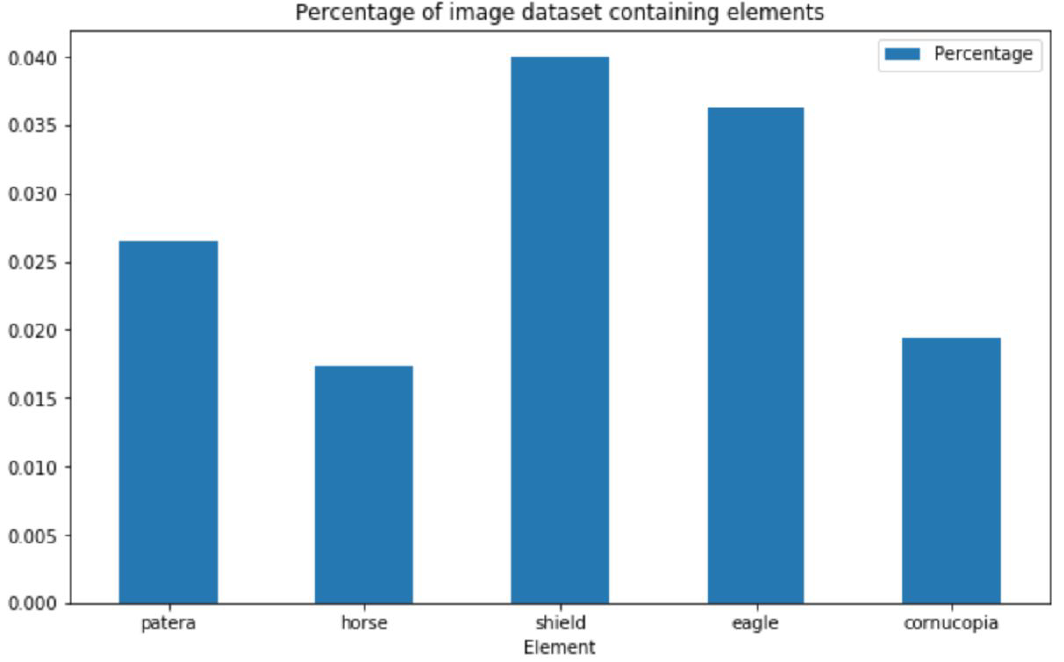}
\caption{Due to the great diversity of different semantic elements which appear on ancient coins only a relatively small number of coins in the corpus as a whole contain the specific semantic elements which we attempt to learn herein, raising class imbalance issues which we address using stratified sampling (see main text for thorough explanation).}
\label{f:expElements}
\end{SCfigure}

In order to select which semantic elements to focus on, we use the unstructured text files to analyse the frequency of word occurrences and, specifically, focus on the most common concepts, see Figs~\ref{f:elements} and~\ref{f:expElements}. Considering that the present work is the first attempt at addressing the problem at hand, the aforementioned decision was made in order to ensure that sufficient training data is available.

\paragraph{Clean-up and normalization of text data}
To label the data we first clean the attribution text files to remove duplicate words and punctuation. Because the data included attributions in French, Spanish, German and English, we employ a translation API (googletrans) to generate translations of each keyword. Thus, when building the data set used to learn the concept of `horse', images with associated text which contains the words `horse', `caballo', `cheval' and `pferd' are also included as positive examples, making better use of available data. Plurals, synonyms and other words strongly associated with the concept in question (e.g.\ `horseman'), and their translations are also included.

\subsubsection{Randomization and stratification} 
We shuffle the samples before building training, validation and test sets for each of the selected elements (`horse', `cornucopia', `patera', `eagle' and `shield'), with a ratio of 70\% training set, 15\% validation set and 15\% test set. To address under-representation of positive examples, we use stratified sampling to ensure equal class representation.

\subsection{Errors in data}
In the context of the present problem as in most large scale, real-world applications of automatic data analysis, even after preprocessing, the resulting data will contain a certain portion of erroneous data. Errors range in nature from incorrect label assignments to incorrectly prepared images. 

To give a specific example, we observed a number of instances in which the keyword used to label whether or not a given coin image contains the element in question was actually referring to the obverse of the coin, rather than the reverse, with which we are concerned. This is most prevalent with shields, and leads to incorrect labelling when building the corresponding data sets. Our premise (later confirmed by empirical evidence) is that such incorrect labelling is not systematic in nature (c.f.\ RANSAC) and thus that the cumulative effect of a relatively small number of incorrect labels will be overwhelmed by the coherence (visual and otherwise) of correctly labelled data. 

We also found that some of the original (unprocessed) images were unusually laid out in terms of the positioning of the corresponding coin's obverse and reverse, which thus end up distorted during preprocessing. However, these appear to be almost exclusively of the auction listing type described above, and as such would be labelled as negative samples and therefore cumulatively relatively unimportant.

\section{Proposed framework}
In this paper we describe a novel deep convolutional neural network loosely based on AlexNet \cite{KrizSutsHint2012} as a means of learning the appearance of semantic elements and thus determining whether an unknown query coin contains the said element or not. Although we do not use this information in the present work for other purposes than the analysis of results, we are also able to determine the location of the element when it is present by generating heatmaps using the occlusion technique \cite{SchlAran2017}.

\subsection{Model topology}
\begin{figure*}[t]
\centering
\includegraphics[width=.9\columnwidth]{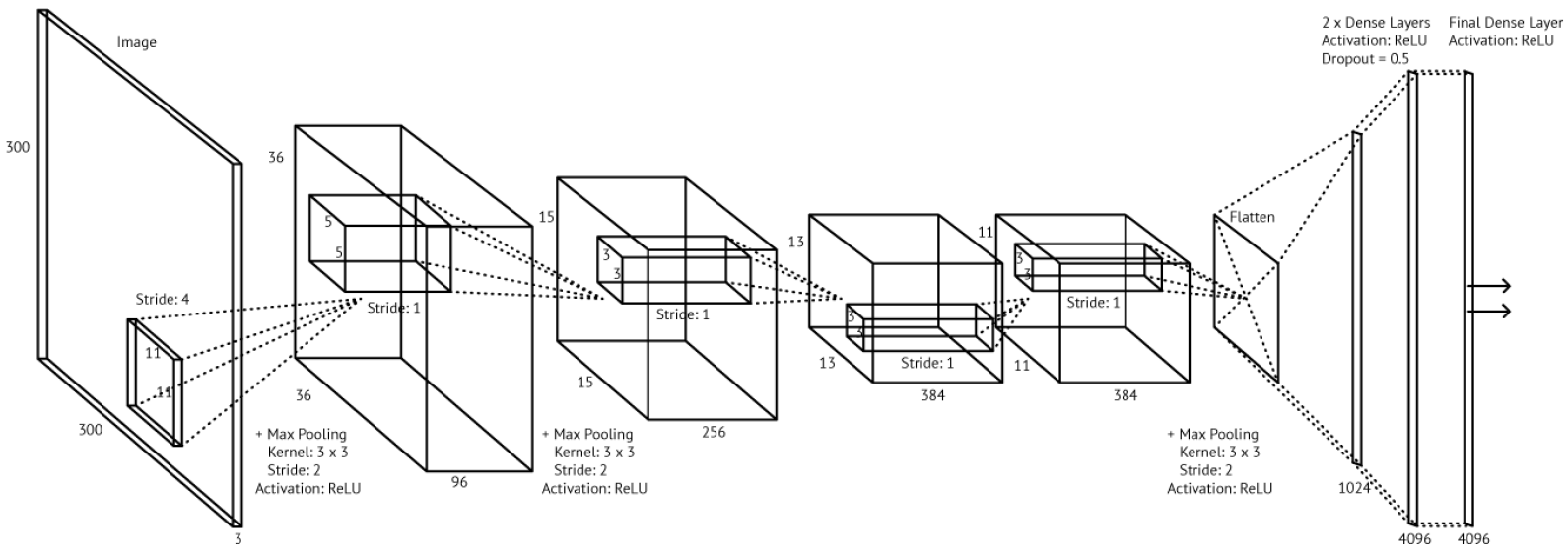}
\caption{Our model architecture: a global view.}
\label{f:arch}
\vspace{-0.10pt}\end{figure*}

As summarized in Fig~\ref{f:arch} and Table~\ref{t:topology}, our network consists of five convolutional layers (three of which are followed by a max pooling layer), after which the output is flattened and passed through three fully connected layers. The convolutional and pooling layers are responsible for detecting high level features in the input data, and the fully connected layers then use those features to generate a class prediction.

The convolutional layers enable location independent recognition of features in the data. The output of a convolutional layer is dependent on a number of hyperparameters which we specify for each layer. These are the stride, the number of kernels (also called the depth), the kernel size, and the padding. Using multiple kernels is necessary in order to learn a range of features. The complexity of the features we are able to detect depends on the number of convolutional layers present in a network -- for example, one can reasonably expect to detect only simple, low-level features such as edges from the original `raw' pixel inputs in the first layer, and then from those edges learn simple shapes in the second layer, and then more complex, higher level features from those simple shapes, and so on.

The pooling layers can be thought of as downsampling -- the aim is to reduce the size of the output whilst retaining the most important information in each kernel. Pooling also decreases the complexity of the network by decreasing the number of parameters we need to learn. It also helps to make the network less sensitive to small variations, translations and distortions in the image input. Our design uses max-pooling, where the highest value in each kernel forms one element in the pooling layer output.

The dense layers (also called fully-connected layers) transform the high level features learned by the previous layers into a class prediction. To avoid overfitting, the dense layers in our model employ dropout, a regularisation technique which combats overfitting by randomly removing nodes at each iteration. The number of nodes that are removed is determined by a hyperparameter which, guided by previous work on CNNs, we set to 0.5.

We use the batch size of 24, and the maximum number of epochs of 200 which we experimentally found to be more than sufficient, with training invariably ending well before that number is reached. At each epoch, we check if the loss is lower than 0.001, or if there have been 30 epochs in a row with no improvement in the loss. If either of the conditions is fulfilled, training is terminated. This is done partly to avoid models overfitting, and partly to save time if a model or a certain set of hyperparameters are clearly not performing well.

The model is trained using adaptive moment estimation (Adam) optimization, a form of gradient descent, the learning rate and momentum being computed for each weight individually \cite{KingAdam2015}. Because we are performing classification (in that each sample either `contains element' or `does not contain element') we use cross entropy loss \cite{Jano2017}. The rectified linear unit (ReLU) \cite{Agar2018} is employed at the activation function to avoid problems associated with vanishing gradients.

\begin{table}[t]
\centering
\setlength{\tabcolsep}{12pt}
\renewcommand{\arraystretch}{1.2}
\caption{Our convolutional neural network topology: a summary.}\vspace{-0.8pt}
\begin{tabular}{lcccc}
\Hline
Layer Type & Kernel size & Stride & Kernel no. & Activation\\
\hline
Convolutional & 11 $\times$ 11 & 4 & 96 & ReLU\\
Max Pooling & 3 $\times$ 3 & 2\\
\hline
Convolutional & 5 $\times$ 5 & 1 & 256 & ReLU\\
Max Pooling & 3 $\times$ 3 & 2\\
\hline
Convolutional & 3 $\times$ 3 & 1 & 384 & ReLU\\
\hline
Convolutional & 3 $\times$ 3 & 1 & 384 & ReLU\\
\hline
Convolutional & 3 $\times$ 3 & 1 & 256 & ReLU\\
Max Pooling Layer & 3 $\times$ 3 & 2\\
\hline
Flatten\\
\hline
& Outputs & Dropout & & \\
\hline
Dense & 4096 & 0.5 & &ReLU\\
\hline
Dense & 4096 & 0.5 & & ReLU\\
\hline
Dense & 2 & 0.5 & & ReLU\\
\Hline
\end{tabular}
\label{t:topology}
\end{table}

\section{Experiments}
Experiments discussed in the present section were performed using Asus H110M-Plus motherboard powered by Intel Core i5 6500 3.2GHz Quad Core processor and NVidia GTX 1060 6GB GPU, 8GB RAM. Data comprised images and the associated textual descriptions of 100,000 auction lots kindly provided to us by the Ancient Coin Search auction aggregator \url{https://www.acsearch.info/}.

\subsection{Results and discussion}
\begin{figure*}[t]
\centering
\includegraphics[width=1\columnwidth]{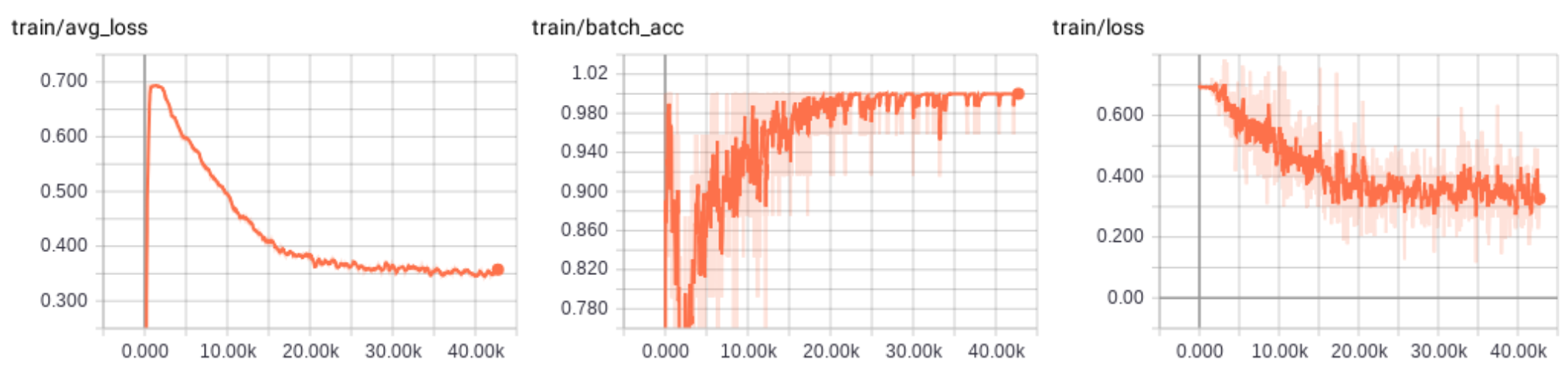}
\caption{Training a model to identify depictions of cornucopiae.}
\label{f:tensorboard}
\end{figure*}

\begin{table}[t]
\centering
\renewcommand{\arraystretch}{1.2}
\setlength{\tabcolsep}{12pt}
\caption{Summary of experimental settings and results.}\vspace{-0.8pt}
\begin{tabular}{l|ccccc}
\Hline
 & Cornucopia & Patera & Shield & Eagle & Horse\\
\hline
Number of epochs & 105 & 136 & 118 & 86 & 148\\
Training time (min) & 58 & 30 & 82 & 51 & 106\\
\hline
Training accuracy & 0.71 & 0.83 & 0.75 & 0.88 & 0.88\\
Validation accuracy & 0.85 & 0.86 & 0.73 & 0.73 & 0.82\\
Validation precision & 0.86 & 0.85 & 0.72 & 0.71 & 0.82\\
Validation recall & 0.83 & 0.86 & 0.75 & 0.81 & 0.84\\
Validation F1 & 0.85 & 0.86 & 0.74 & 0.75 & 0.83\\
\hline
\bf Test accuracy & 0.84 & 0.84 & 0.72 & 0.73 & 0.82\\
\bf Test precision & 0.85 & 0.82 & 0.71 & 0.70 & 0.81\\
\bf Test recall & 0.83 & 0.87 & 0.74 & 0.81 & 0.82\\
\bf Test F1 & 0.84 & 0.84 & 0.72 & 0.75 & 0.82\\
\Hline
\end{tabular}
\label{t:results}
\end{table}

\begin{figure*}[t]
  \centering
  \includegraphics[width=0.25\columnwidth]{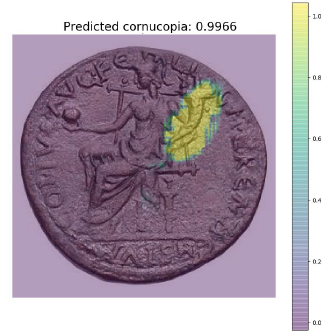}\hspace{30pt}
  \includegraphics[width=0.25\columnwidth]{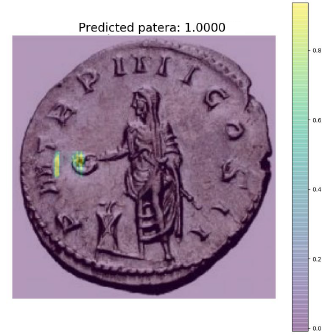}\hspace{30pt}
  \includegraphics[width=0.25\columnwidth]{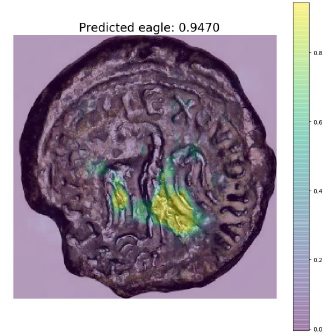}
\caption{Examples of automatically identified salient regions corresponding to a cornucopia, a patera, and a shield, respectively.}
\label{f:heatmaps}
\vspace{-0.10pt}\end{figure*}

As readily seen from Table~\ref{t:results}, our approach achieves a high level of accuracy across training, validation, and test data. The validation and test scores are not significantly lower than the training scores, which suggests little evidence of overfit and indicates a well designed architecture and sufficient data set. The models used to identify cornucopiae and paterae are particularly successful, which is likely due to the low level of variance in artistic depictions of these elements -- the orientation, position and style of shields, horses and eagles varies quite a bit, but cornucopiae and paterae are fairly constant in their depiction.

\subsection{Learnt salient regions}
We use the occlusion technique \cite{SchlAran2017} to quantify the importance of different image regions in the context of the specific task at hand. In brief, the process involves synthetic occlusion by a uniform kernel and the quantification of the corresponding classification performance differential between unoccluded and occluded inputs, with a higher difference suggesting higher importance. Previous work on computer vision based ancient coin analysis has demonstrated the usefulness of this technique in the interpretation of empirical results \cite{SchlAran2017}. In order to ensure robustness to relative size (semantic element to coin diameter), in the present work we adopt the use of three kernel sizes, $32 \times 32$, $48 \times 48$, and $64 \times 64$ pixels.

Typical examples of identified salient regions for different semantic elements are shown in Fig~\ref{f:heatmaps}. It can be readily seen that our algorithm picks up the most characteristic regions of the elements well -- the winding pattern of cornucopiae, the elliptical shape of paterae, and the feather patterns of eagles. Since, as we noted before and as is apparent from Table~\ref{t:results}, the performance of our algorithm is somewhat worse on the task of shield detection, we used our occlusion technique to examine the results visually in more detail. Having done so, our conclusion is reassuring -- our hypothesis is that shields are inherently more challenging to detect as they exhibit significant variability in appearance and the style of depiction (shown from the front they appear circular, whereas shown from the side they assume an arc-like shape) and the least amount of characteristic detail (both circular and arc-like shapes are commonly found in more complex semantic elements shown on coins).

\section{Summary and conclusions}
In this paper we made a series of important contributions to the field of computer vision based analysis of ancient coins. Firstly, we put forward the first argument against the use of visual matching of ancient coin images, having explained its practical lack of value. Instead, we argued that efforts should be directed towards the semantic understanding of coin images and described the first attempt at this challenging task. Specifically, we described a novel approach which combines unstructured text analysis and visual learning using a convolutional neural network, to create weak associations between semantic elements found on ancient coins and the corresponding images, and hence learn the appearance of the aforementioned elements. We demonstrated the effectiveness of the proposed approach using images of coins extracted from 100,000 auction lots, making the experiment the largest in the existing literature. In addition to a comprehensive statistical analysis, we presented the visualization of learnt concepts on specific instances of coins, showing that our algorithm is indeed creating the correct associations. We trust that our contributions will serve to direct future work and open avenues for new promising research.

\section*{Acknowledgements}
The authors gratefully acknowledge the support of NVIDIA Corporation for their donation of
the GPU used for this research.

\tiny
\bibliographystyle{splncs}
\bibliography{oa_bibliography}

\begin{thebibliography}{10}
\providecommand{\url}[1]{\texttt{#1}}
\providecommand{\urlprefix}{URL }
\providecommand{\doi}[1]{https://doi.org/#1}

\bibitem{Agar2018}
Agarap, A.F.: Deep learning using rectified linear units ({ReLU}). arXiv p.
  1803.08375 (2018)

\bibitem{AnwaZambKamp2013}
Anwar, H., Zambanini, S., Kampel, M.: Supporting ancient coin classification by
  image-based reverse side symbol recognition. In Proc.\ International
  Conference on Computer Analysis of Images and Patterns pp. 17--25 (2013)

\bibitem{AnwaZambKamp2015}
Anwar, H., Zambanini, S., Kampel, M.: Coarse-grained ancient coin
  classification using image-based reverse side motif recognition. Machine
  Vision and Applications  \textbf{26}(2),  295--304 (2015)

\bibitem{Aran2010}
Arandjelovi{\'c}, O.: Automatic attribution of ancient {R}oman imperial coins.
  In Proc.\ IEEE Conference on Computer Vision and Pattern Recognition pp.
  1728--1734 (2010)

\bibitem{Aran2012d}
Arandjelovi{\'c}, O.: Reading ancient coins: automatically identifying denarii
  using obverse legend seeded retrieval. In Proc.\ European Conference on
  Computer Vision  \textbf{4},  317--330 (2012)

\bibitem{ConnAran2017}
Conn, B., Arandjelovi{\'c}, O.: Towards computer vision based ancient coin
  recognition in the wild -- automatic reliable image preprocessing and
  normalization. In Proc.\ IEEE International Joint Conference on Neural
  Networks pp. 1457--1464 (2017)

\bibitem{FareAran2017}
Fare, C., Arandjelovi{\'c}, O.: Ancient {R}oman coin retrieval: a new dataset
  and a systematic examination of the effects of coin grade. In Proc.\ European
  Conference on Information Retrieval pp. 410--423 (2017)

\bibitem{Jano2017}
Janocha, K., Czarnecki, W.M.: On loss functions for deep neural networks in
  classification. arXiv p. 1702.05659 (2017)

\bibitem{KampZaha2008}
Kampel, M., Zaharieva, M.: Recognizing ancient coins based on local features.
  In Proc.\ International Symposium on Visual Computing  \textbf{1},  11--22
  (2008)

\bibitem{KingAdam2015}
Kinga, D., Adam, J.B.: A method for stochastic optimization. In Proc.\
  International Conference on Learning Representations  \textbf{5} (2015)

\bibitem{KrizSutsHint2012}
Krizhevsky, A., Sutskever, I., Hinton, G.E.: Imagenet classification with deep
  convolutional neural networks. Advances in Neural Information Processing
  Systems pp. 1097--1105 (2012)

\bibitem{Matt1966}
Mattingly, H.: The {R}oman {I}mperial coinage., vol.~7. Spink (1966)

\bibitem{RieuAran2015}
Rieutort-Louis, W., Arandjelovi{\'c}, O.: {Bo(V)W} models for object
  recognition from video. In Proc.\ International Conference on Systems,
  Signals and Image Processing pp. 89--92 (2015)

\bibitem{RieuAran2016}
Rieutort-Louis, W., Arandjelovi{\'c}, O.: Description transition tables for
  object retrieval using unconstrained cluttered video acquired using a
  consumer level handheld mobile device. In Proc.\ IEEE International Joint
  Conference on Neural Networks pp. 3030--3037 (2016)

\bibitem{SchlAran2017}
Schlag, I., Arandjelovi{\'c}, O.: Ancient {R}oman coin recognition in the wild
  using deep learning based recognition of artistically depicted face profiles.
  In Proc.\ IEEE International Conference on Computer Vision pp. 2898--2906
  (2017)

\bibitem{YueDimiAran2019}
Yue, X., Dimitriou, N., Arandjelovi{\'c}, O.: Colorectal cancer outcome
  prediction from {H\&E} whole slide images using machine learning and
  automatically inferred phenotype profiles. In Proc.\ International Conference
  on Bioinformatics and Computational Biology  (2019)

\bibitem{ZahaKampZamb2007}
Zaharieva, M., Kampel, M., Zambanini, S.: Image based recognition of ancient
  coins. In Proc.\ International Conference on Computer Analysis of Images and
  Patterns pp. 547�--554 (2007)

\end{thebibliography}

\end{document}